\PassOptionsToPackage{prologue,dvipsnames}{xcolor}
\documentclass[final]{l4dc2026/l4dc2026}
\usepackage{amsfonts,amssymb}
\usepackage{bbm}

\usepackage{algorithm}
\usepackage{algpseudocode}
\usepackage{algorithmicx}
\usepackage{sidecap}

\usepackage[dvipsnames]{xcolor}
\usepackage[normalem]{ulem}
\usepackage{hhline}

\usepackage{booktabs}
\usepackage{titlesec}
\usepackage{fancyhdr}

\usepackage{enumitem}
\setlist{noitemsep} 

\usepackage{graphicx}
\usepackage{capt-of} 
\usepackage{multirow}

\usepackage{hyperref}
\hypersetup{colorlinks=true,linkcolor=magenta,citecolor=gray,urlcolor=gray,pagebackref=true}

\usepackage{endnotes}

\newtheorem{thm}{Theorem}

\newtheorem{defi}{Definition}

\newcommand{\bla}{\begin{block}}
\newcommand{\blb}{\end{block}}

\newcommand{\defa}{\begin{defi}}
\newcommand{\defb}{\end{defi}}
\usepackage{hyperref}

\newcommand{\thma}{\begin{thm}}
\newcommand{\thmb}{\end{thm}}

\newcommand{\mata}{\begin{bmatrix}}
\newcommand{\matb}{\end{bmatrix}}

\newcommand{\argmax}{\operatornamewithlimits{argmax}}
\newcommand{\argmin}{\operatornamewithlimits{argmin}}

\newcommand{\TT}{^{\ensuremath{\mathsf{T}}}}           
\providecommand{\mc}[1]{\mathcal{#1}}
\providecommand{\mb}[1]{\boldsymbol{#1}}

\newcommand{\Real}{\mathbb{R}}

\newcommand{\bX}{\mb{X}}
\newcommand{\bY}{\mb{Y}}

\newcommand{\DD}{\mathbb{D}}

\newcommand{\HH}{\mathbb{H}}
\newcommand{\LL}{\mathbb{L}}

\newcommand{\XX}{\mathbb{X}}
\newcommand{\YY}{\mathbb{Y}}
\newcommand{\ZZ}{\mathbb{Z}}

\newcommand{\eps}{\varepsilon}

\usepackage{subcaption}
\usepackage{cleveref}
\Crefname{equation}{Eq.}{Eqs.} 

\usepackage{amsfonts,bm,dsfont,bbm, amssymb}
\usepackage{scalerel}

\def\eqref#1{equation~\ref{#1}}

\def\1{\bm{1}}

\def\eps{{\epsilon}}

\DeclareMathAlphabet{\mathsfit}{\encodingdefault}{\sfdefault}{m}{sl}
\SetMathAlphabet{\mathsfit}{bold}{\encodingdefault}{\sfdefault}{bx}{n}

\providecommand{\mc}[1]{\mathcal{#1}}

\let\P\relax
\DeclareMathOperator{\P}{\mathbb{P}}
\DeclareMathOperator*{\E}{\mathbb{E}}

\renewcommand{\implies}{\Rightarrow}

\def \s {\sigma}

\def \TT {\mathcal{T}}

\def \XX {\mathcal{X}}
\def \YY {\mathcal{Y}}
\def \LL {\mathcal{L}}

\def \DD {\mathcal{D}}

\def \HH {\mathcal{H}}

\Crefname{example}{Example}{Examples}
\Crefname{scenario}{Scenario}{Scenarios}

\providecommand{\mc}[1]{\mathcal{#1}}

\newlist{todolist}{itemize}{2}
\setlist[todolist]{label=\cross}



\hypersetup{
    colorlinks=true,
    linkcolor=cyan!70!black,
    filecolor=magenta,      
    urlcolor=cyan,
    citecolor=cyan!60!black,
}

\def \s {\sigma}

\def \TT {\mathcal{T}}

\def \XX {\mathcal{X}}
\def \YY {\mathcal{Y}}
\def \LL {\mathcal{L}}

\def \DD {\mathcal{D}}

\def \HH {\mathcal{H}}

\def \naturals {\mathbb{N}}
\def \Reals {\mathbb{R}}

\usepackage{lipsum}

\usepackage{xcolor}
\definecolor{commentgreen}{RGB}{34,139,34} 

\title{Optimal Control of the Future via \\
~Prospective Learning with Control}
\usepackage{times}
\author{%
 \Name{Yuxin Bai}\nametag{$^{1}$} \Email{ybai31@jh.edu}
 \AND 
 \Name{Aranyak Acharyya}\nametag{$^{2} \thanks{Equal contribution}$} \Email{aachary6@jh.edu}
 \AND 
 \Name{Ashwin {De Silva}}\nametag{$^{1*}$}
 \Email{ldesilv2@jh.edu}
 \AND 
 \Name{Zeyu Shen}\nametag{$^{3}$}
 \Email{zshen39@jh.edu}
 \AND
 \Name{James Hassett}\nametag{$^{1}$}
 \Email{jhasset1@jh.edu}
 \AND
 \Name{Joshua T.~Vogelstein}\nametag{$^{1}$}
 \Email{jovo@jh.edu}\\
\addr $^{1}$ Department of Biomedical Engineering, Johns Hopkins University, Baltimore, MD 21218\\
\addr $^{2}$ Mathematical Institute for Data Science, Johns Hopkins University, Baltimore, MD 21218\\
\addr $^{3}$ Department of Applied Mathematics and Statistics, Johns Hopkins University, Baltimore, MD 21218
}

\begin{document}

\maketitle

\begin{abstract}
Optimal control of the future is the next frontier for AI. Current approaches to this problem are typically rooted in reinforcement learning (RL).  RL is mathematically distinct from supervised learning, which has been the main workhorse for the recent achievements in AI. Moreover, RL typically operates in a stationary environment with episodic resets, limiting its utility. Here, we extend supervised learning to address learning to \textit{control} in non-stationary, reset-free environments. Using this framework, called ''Prospective Learning with Control'' (PLuC), we prove that under certain fairly general assumptions, empirical risk minimization (ERM) asymptotically achieves the Bayes optimal policy.  We then consider a specific instance of prospective learning with control: foraging, a canonical task relevant to both natural and artificial agents. We illustrate that modern RL algorithms, which assume stationarity, struggle in these non-stationary reset-free environments. Even with time-aware modifications, they converge orders of magnitude slower than our prospective foraging agents on a simple 1-D foraging benchmark.  
Code is available at: \url{https://github.com/neurodata/procontrol}.

\end{abstract}


\section{Introduction}
When building artificial intelligence (AI) systems, we often start with some basic operating assumptions about the world.  For example, a prevailing, though clearly limited assumption, is that the data-generating process is stationary.  This is a core assumption in much of foundational estimation and learning theory dating back nearly a century~\cite{Glivenko1933-wm, Cantelli1933-vj, Vapnik1971-um, Valiant1984-dx}, all the way through modern machine learning tools including large language models~\cite{vaswani2017attention}.  
A second prevailing and obviously wildly simplifying assumption is that the learner's decisions have no impact on the data distribution. This assumption leads to developing systems based purely on predictions and forecasts, without concerning oneself with the consequences of one's actions. 
For example, consider early machine learning successes such as spam detection~\cite{Sahami1998-yr}, which operated according to both assumptions: no time and no impact.  Once it is deployed, spammers change their behavior to subvert the system.  This simple example illustrates a few key properties required of deployed machine learning systems.

First, time is real.  In particular, the world is dynamic; therefore, for the system to preserve its desirable properties, it must also be dynamic on approximately the same time-scales.  Second, decisions have impacts.  Therefore, pure inference or forecasting systems are missing a key ingredient. 
The history of machine learning has traversed a trajectory of starting with the simplest possible assumptions, and slowly relaxing these assumptions to obtain more effective deployed systems. One limitation of relaxing the simplifying assumptions is that as we strive to estimate more complex functions, we require more data.  Thus, as our ability to acquire and analyze more high-quality data has improved, so has our toolbox generalized accordingly.  

Nearly one hundred years ago, people realized that one could estimate arbitrary properties of data under quite general assumptions, as long as we assumed no time and no impact~\cite{Glivenko1933-wm, Cantelli1933-vj}. In the 1950s, decision theory seriously relaxed the assumption of no impact~\cite{Wald1949-fl}. Markov models and Hidden Markov Models (HMM) relaxed the assumption of no time~\cite{Baum1966-ep}, and Markov decision processes (MDP) and Partially Observable MDPs (POMDPs) further relaxed the assumption of no impact with the Bellman equation~\cite{Bellman1958-us}, introducing control theory. Then Kalman filters addressed the fact that time and impact were real, but the parameters of the system were partially unknown~\cite{Kalman1960-mm}, founding adaptive control~\cite{Landau1984-mr}.  Modern approaches to adaptive control are typically considered reinforcement learning (RL)~\cite{Sutton2018-gx}.  
Modern ML has many successes, though they are typically restricted to settings where the environment is stationary, there are many trials, and the whole world is perfectly observable (e.g., in video games~\cite{Silver2016NatureGo}).  Non-stationary adaptive control and RL, and continual RL address the possibility of an environment that changes over time~\cite{Khetarpal2022-tc, Abel2023-gl, Kumar2025-go}.  However, they typically operate under the assumption of many trials, rather than single-life~\cite{Chen2022-dq} and reset-free~\cite{Lu2020-bp} approaches.  As is the norm in RL, these approaches continue to operate under a Markov Decision Process framework. 

Our contribution builds on these earlier works, proposing a framework for non-stationary reset-free environments that does not require the standard MDP assumption.  In a departure from much of RL, our work builds on estimation theory~\cite{Glivenko1933-wm, Cantelli1933-vj},  Probably Approximately Correct learning theory~\cite{Valiant1984-dx, Valiant2013-gp}, and modern machine learning and supervised learning practice~\cite{Mohri2012-qj,Goodfellow2016-kf}.  Continual learning~\cite{Vogelstein2025-uq} and prospective learning~\cite{De_Silva2023-al, De-Silva2024-jj,Bai2026-ik} generalize estimation and learning theory by relaxing the assumption of no time.  In this work, we introduce Prospective Learning with Control, which is denoted by \textbf{P}rospective \textbf{L}earning pl\textbf{u}s \textbf{C}ontrol (\textbf{PLuC}), where we further relax the assumption of no impact.

We formally introduce our framework in~\Cref{sec:foraging}. 
Algorithms for solving PLuC problems are provided in~\Cref{sec:proforg}.  Numerical results are provided in~\Cref{sec:exp_val}, demonstrating that leading RL algorithms, even augmented to incorporate non-stationarity, either fail to converge to the Bayes optimal, or converge orders of magnitude more slowly than our algorithms.  Theory proving that our algorithms converge to the Bayes optimal solution under suitable assumptions is in~\Cref{sec:Theoretical_Results}. 
This initial collection of empirical and theoretical results suggests that the prospective learning with control is a promising direction worthy of further study, particularly as it is extended to more complex domains. 


\section{Framework for Prospective Learning with Control (PLuC)} 
\label{sec:foraging}

This work builds on prospective learning defined previously~\cite{De_Silva2023-al, De-Silva2024-jj, Bai2026-ik}, which we refer to hereafter as Prospective Learning without Control (PLiC).  See~\Cref{sec:dynamic} for a review of that work with slightly modified notation,  and contrast with Prospective Learning with Control (PLuC), as defined below. 


\paragraph{Data}  Let $x_t$ be the position of the agent in an environment at time $t$. 
%
We also record time $t \in \mc{T} \subset \naturals \equiv [1,2,3,\dots]$. We refer to the position and time together as the observation state (or simply state),  $(x_t,t)$. 
Let $y_t \in \YY$ be the reward the agent would receive at time $t$ at any of the positions. We define the observed datum at time $t$ as $Z_t = (X_t, Y_t)$, with realization $z_t = (x_t, y_t)$.We denote the observed history up to time $t$ by $z_{\le t} = (z_1,\dots,z_t)$, and the unobserved future data by $z_{>t}$. We model the sequential interaction between the agent and environment as a stochastic decision process. In particular, the observed sequence $Z = (Z_t)_{t\in\mathbb{N}}$ is a stochastic process defined on a probability space $(\Omega,\mathcal{F},\mathbb{P})$.

\paragraph{Hypothesis} The hypothesis space generalizes the notion from PLiC in ~\Cref{sec:dynamic}. Here, $h_t:\XX \times \TT  \to \XX \times \YY$ is a hypothesis in $\mc{H}_t$, where $\mc{H}_t \subseteq  \{h: \XX \times \TT \to \XX \times \YY\}$, is a hypothesis class characterizing the environment's dynamics and feedback. Therefore, $h_t$ infers an output $(x_{t+1},y_{t+1})$ for a given input $x_t$ at time $t$.%


\paragraph{Loss}
The instantaneous loss is the negative reward the agent receives at the position the agent is in at that time. Recalling that 
$x_s$ encodes the  position of the agent at time $s$, and that $y_{s+1}$ encodes the reward at all locations at time $s+1$, 
we can write
\begin{equation} \label{eq:PFloss}
\bar{\ell}(h_s(x_s,s),y_{s+1}) =  - y_{s+1}(x_{s+1}).  
\end{equation}
%
The loss at time $t$ is the weighted cumulative instantaneous loss\footnote{We intentionally do not specify whether the horizon is finite or infinite here.},
$$\ell_t(h, Z) = \sum_{s>t} w_{s-t} \bar{\ell} (h_s(x_s,s), y_{s+1}).$$

\paragraph{Learner}
The learner $\LL$ is a map from the observed data, $\DD_t$, to a hypothesis, $\hat{h}_t$.  
The observed data are 
$\DD_t=(x_s,y_{s+1}(x_{s+1}),s)_{s \leq t-1}$.


\paragraph{Theoretical Goal} 
We define risk as the expected loss: 
$$R_t(h) = \mathbb{E}_F \left[ \ell_t(h, Z) \mid z_{\leq t} \right]  = \int \ell_t (h, Z ) \ \mathrm{d} \mathbb{P}_{Z \mid z_{\leq t}}.$$
%
 %
%
%
The goal is to choose a learner that obtains a hypothesis that minimizes risk. 
 

\begin{equation} \label{eq:PFgoal}
\hat{h}^t =  \argmin_{h \in \mc{H}_t} R_t(h)  
\equiv \argmax_{h \in \mc{H}_t}  \int \sum_{s>t} w_{s-t} \cdot y_{s+1}^h(x_{s+1}^h) \ \mathrm{d} \mathbb{P}_{Z \mid z_{\leq t}},
\end{equation}
where $y^h_{s+1}$ and $x^h_{s+1}$ indicate that the subsequent positions and rewards are functions of $h$.  

\paragraph{Empirical Goal}
We cannot directly solve~\Cref{eq:PFgoal} because 
we cannot sample all possible futures, and we do not know the likelihood of each possible future.  
We therefore replace the integral in~\Cref{eq:PFgoal} with a sum up to the current time, $t$:

\begin{equation}    \label{eq:perm2}
\hat{h}^t  
\approx \argmin_{h^t \in \mc{H}_t} \sum_{t' > 0 }^t \sum_{s> t'}^t - w_{s-t'} \cdot y_{s+1}(x_{s+1}).  
\end{equation}

\section{Theory for Prospective Learning with Control}
\label{sec:Theoretical_Results}
Here we provide proof sketches under asymptotic assumptions; formal proofs are in~\Cref{s:proofs}.
We note that finite-sample convergence rates remain an open problem.

\begin{lemma}
\label{Lm:expected_sup_partial_sum_to_Bayes_risk}
    Suppose $\lbrace Z_t \rbrace_{t=1}^{\infty}$ is a stochastic process and let $\lbrace \mathcal{H}_t \rbrace_{t=1}^{\infty}$ be an increasing hypothesis class, such that there exists $h_t \in \mathcal{H}_t$ which satisfies
$        \lim_{t \to \infty}
        \mathbb{E}[R_t(h_t)-R_t^*]=0.
$    
    Additionally, suppose $\lbrace u_t \rbrace_{t=1}^{\infty}$ is  a sequence such that $u_t \to \infty$ as $t \to \infty$, and $u_t \leq t$ for all $t$. Define the partial cumulative prospective loss to be
        $e_m(h)=
        \sum_{s=1}^m w_{s} \tilde{y}_{s+1}(x_{s+1}).~$
    Then, there exists $\tilde{h}_{t} \in \mathcal{H}_t$ such that
$    \lim_{t \to \infty}
        \mathbb{E}
        \left[
\sup_{u_t \leq m \leq \infty}
e_m(\tilde{h}_{t})|Z_{\leq t}
        \right]-R^*_t =0.
$    
\end{lemma}

\paragraph{Sketch of proof} A condition for the above lemma to hold is the existence of a sequence of hypotheses whose risks converge to the Bayes optimal risk. We establish that for this asymptotically optimal sequence of hypotheses, there exists a subsequence of timepoints on which the expected partial cumulative loss is arbitrarily close to the Bayes risk, using Markov's Inequality and Borel-Cantelli Lemma. We use this subsequence of timepoints to establish that for a particular sequence of hypotheses, the expected partial cumulative loss approaches the Bayes optimal risk.

\begin{theorem}
\label{Thm:consistency_and_concentration_implies_approaching_Bayes_risk}
        Consider a family of stochastic processes $(\mathbf{X}_t,\mathbf{Y}_t)_{t>0}$. Suppose there exists a hypothesis class such that $\mathcal{H}_1 \subseteq \mathcal{H}_2 \dots  $ such that 
$        \lim_{t \to \infty}
        \left[
\inf_{h \in \mathcal{H}_t} R_t(h)-R^*_t
        \right]=0. 
$    
Also, there exist sequences $\lbrace u_t \rbrace_{t=1}^{\infty}$ and $\lbrace \xi_t \rbrace_{t=1}^{\infty}$ satisfying $\lim_{t \to \infty} \xi_t=0$, and $u_t \leq t$, such that

\begin{equation*}
    \mathbb{E}
    \left[
\max_{h \in \mathcal{H}_t}
\left|
\sum_{s=t+1}^{\infty}
w_{s-t} \tilde{y}_{s+1}(x_{s+1}) -
\max_{u_t \leq m \leq t}
\sum_{s=1}^{m} w_{s} \tilde{y}_{s+1}(x_{s+1})
\right|
    \right] \leq \xi_t.
\end{equation*}
\vspace{0.1cm}
Then, there exists a sequence $\lbrace i_t \rbrace_{t=1}^{\infty}$ such that 

$    \hat{h}^{(t)}=
    \argmin_{h \in \mathcal{H}_{i_t}} \max_{u_{i_t} \leq m \leq t}
    \sum_{s=1}^m 
    w_{s} \tilde{y}_{s+1}(x_{s+1})
~$
reaches the Bayes optimal risk.
\end{theorem}
\paragraph{Sketch of proof} We note that for any sequence of hypotheses, the difference between the prospective loss and the partial cumulative loss is bounded by terms approaching zero. We find a subsequence of timepoints on which this difference is sufficiently small, and using Lemma \ref{Lm:expected_sup_partial_sum_to_Bayes_risk} we establish that the Bayes risk of the minimizer of the partial cumulative loss on that particular subsequence of timepoints approaches zero.     

\section{Algorithms for Prospective Learning with Control}
\label{sec:proforg}
\label{s:algorithm_l4dc}
\begin{algorithm}[H]
\caption{Prospective Learner with Control (PLuC)}
\begin{algorithmic}[1]

\State \textbf{Initialize:} Instantaneous loss model $g_i(x_t, t \ ; \theta)$, Cumulative loss model $g_i(x_t, t \ ; \varphi)$, Planning horizon $H$, Discount factor $\gamma$, Replay buffer $\mathcal{D}$, Initial position $x_0$, sampling parameters $\{\eta_t\}_{t=0}^{\infty}$
\Statex \textcolor{commentgreen}{$\triangleright$ \texttt{Warm-up phase using \eqref{eq:warmup}}}
\State \textbf{for} $t = 0, 1, 2, \dots, T_{\text{warmup}}-1$
    \State \qquad Randomly sample and move to the next allowed position $x_t$ 
    \State \qquad Observe the instantaneous losses $y_{t}(x_t)$ and store $(x_t, t, y_{t}(x_t))$ in $\mathcal{D}$
\State \textbf{end for}
\Statex
\vspace{0.2cm} \textcolor{commentgreen}{$\triangleright$ \texttt{Online learning and planning phase}}
\State \textbf{for} $t = T_{\text{warmup}} \dots, T_{\text{terminal}}$
    \Statex \qquad \textcolor{commentgreen}{$\triangleright$ \texttt{Update the models}}
    \State \qquad Calculate cumulative losses $\sum_{k=t+1}^T \gamma^{k - t - 1} y_k$
    \State \qquad Update $g_i$ to minimize MSE on observed instantaneous losses
    \State \qquad Update $g_c$ to minimize MSE on calculated cumulative losses
    \vspace{0.2cm}
    \Statex \qquad \textcolor{commentgreen}{$\triangleright$ \texttt{Selecting next position to move according to \eqref{eq:erm}}}
    \State \qquad Identify all possible position sequences $\boldsymbol{x}_{t:t+H}$
        \State \qquad \textbf{for} each sequence $\boldsymbol{x}_{t:t+H}$
            \State \qquad \qquad Calculate the finite horizon loss $Q_{\text{finite}} = \sum_{h=1}^{H} \gamma^{h-1} g_i(x_{t+h}, t+h)$
            \State \qquad \qquad Calculate the terminal loss $Q_{\text{terminal}} = \gamma^H \cdot g_c(x_{t+H}, t+H)$
            \State \qquad \qquad Total sequence loss $Q(\boldsymbol{x}_{t:t+H}) = Q_{\text{finite}} + Q_{\text{terminal}}$
        \State \qquad \textbf{end for}
    \vspace{0.2cm} 
    \Statex \qquad \textcolor{commentgreen}{$\triangleright$ \texttt{Learner-environment interaction by \eqref{sampling}}}
        \State \qquad Construct a sampling distribution over trajectories:
    $p_t(\boldsymbol{x}_{t:t+H}) \propto \exp \left(\eta_t \cdot Q(\boldsymbol{x}_{t:t+H}) \right)$
    \State \qquad Sample the selected trajectory from this distribution: $\boldsymbol{x}^* \sim p_t$
    
    \State \qquad Move to the first position of the optimal sequence $x_t = x^*_t$
    \State \qquad Observe the instantaneous loss $y_{t+1}(x_t)$ and store $(x_t, t, y_{t+1}(x_t))$ in $\mathcal{D}$
    
\State \textbf{end for}

\end{algorithmic}
\end{algorithm}

Inspired by online re-planning strategies in RL~\citep{Bertsekas2023}, we adopt an online execution framework. Prior to iterative policy updates, we perform a warm-up phase during $t = 0, \dots, T_{\text{warmup}}$ where the agent follows a nearly random policy to collect initial interaction data (states and rewards). This pretraining step alleviates the ``cold-start'' problem and facilitates more effective exploration in the early stages of the sequence. Following this phase, at each online iteration, we estimate both cumulative and instantaneous losses to select the next action based on the current learners.

\subsection{Warm Start via Batch Pretraining}

\paragraph{Cumulative loss estimation}
Recall the empirical objective for prospective learning with control from \Cref{eq:perm2}. To operationalize this, we construct an estimator for the prospective loss. We have the states, times, and the corresponding rewards from the warm-start interaction data. We construct a dataset with inputs $\{(\varphi_x(x_s), \varphi_t(s))\}_{s \leq T_{\text{warmup}}}$ and targets consisting of the cumulative rewards:
\begin{align} 
\label{eq:warmup}
Y_s = \sum_{t' = s+1}^{T_{\text{warmup}}} -w_{t'-s} \cdot y_{t'}(x_{t'})
\end{align}

Here, $\varphi_x(\cdot)$ is an state-encoding (such as one-hot encoding) and $\varphi_t(\cdot)$ is a time-encoding comprising of sinusoids and cosines of different frequencies. We then train a supervised regressor $\hat g_c$ —such as a deep neural network or a decision forest—that learns to map these state-time pairs to an estimated cumulative loss. This estimated cumulative loss serves as the prospective loss at state $x_s$ and time $s$, providing the agent with a heuristic for the long-term consequences of its current position.

\paragraph{Instantaneous loss estimation}
When evaluating various hypotheses, many will propose trajectories $x'_s$ that deviate from the historical positions $x_s$ selected by the previous hypothesis $\hat{h}^s$. These counterfactual positions lack direct observations of $y_s$. Consequently, \Cref{eq:perm2} cannot be solved directly due to missing data. To address this, we introduce an approximation $\tilde{y}_{s+1}$. Let $x'_s$ denote a potential outcome (the position the agent could take) and $x_s$ denote the actual observed position. We define the estimated reward $\tilde{y}_s$ as:
\begin{align} 
\label{eq:counter}
\tilde{y}_s(x', s) =
\begin{cases}
y_s(x_s) & \text{if } x'_{s} = x_s \\
\hat{y}_s(x'_s) & \text{if } x'_s \neq x_s  \qquad \triangleright \text{these are counterfactuals} 
\end{cases}
\end{align}
where $\hat{y}_s(x'_s)$ is the reward predicted by an estimator for a counterfactual state. Incorporating this into our objective yields:
$$ \hat{h}^t \approx \arg\min_{h^t \in \mathcal{H}_t} \sum_{t' > 0 }^t \sum_{s> t'}^t - w_{s-t'} \cdot \tilde{y}_{s+1}(x_{s+1}). $$
By utilizing these estimators, we are no longer constrained by the current time $t$. Given a predefined horizon $H$, we can estimate the loss over future intervals:
$$ \hat{h}^t \approx \arg\min_{h^t \in \mathcal{H}_t} \sum_{t' > 0 }^t \sum_{s> t'}^{t+H} - w_{s-t'} \cdot \tilde{y}_{s+1}(x_{s+1}). $$ 
To construct the estimator $\hat{y}_s(\cdot)$, we train a supervised regressor $\hat{g}^i$ on inputs $\{(\varphi_x(x_s), \varphi_t(s))\}_{s \leq T_{\text{warmup}}}$ and targets $\{y_s(x_s)\}_{s \leq T_{\text{warmup}}}$.

\subsection{Online estimation and inference}

Our framework combines estimated instantaneous losses and cumulative losses to approximate the infinite-horizon prospective loss (recall that our goal is~\Cref{eq:perm2}). This approach provides counterfactual insight while maintaining a tractable approximation of future costs. For a given horizon $H$, we estimate the hypothesis as:
\begin{align}
\label{eq:erm}
 \hat{h}^t \approx \arg\min_{h^t \in \mathcal{H}_t} \sum_{t' > 0 }^t \left\{ -w_{t+H-t'} \cdot \bar{y}_{t+H} + \sum_{s> t'}^{t+H} - w_{s-t'} \cdot \tilde{y}_{s+1}(x_{s+1}) \right\} 
\end{align}
where $\bar{y}_{t+T}$ serves as the terminal cost estimation~\citep{Bertsekas2023} at the horizon boundary, and $\tilde{y}_s$ is the instantaneous cost. This structure mirrors ``truncated rollout with terminal cost approximation'', employing two models: one for short-term look-ahead and another to account for the truncated tail of the horizon. 

\paragraph{Inference in Practice}
We evaluate potential future trajectories $\boldsymbol{x}_{s:s+H}$ by calculating their total prospective loss:
$$ \hat{Q}(\boldsymbol{x}_{s:s+H}) = - \left( w_T \cdot \hat{g}^c(x_{s+T}, s+T) + \sum_{k=s}^{s+T} w_{k-s} \cdot \hat{g}^i(x_k, k) \right). $$
For example, with a horizon $T=6$ and action space $A=3$, there are $3^6=729$ possible trajectories. We select the next state stochastically using a Boltzmann rule with a decaying inverse-temperature parameter $\eta_s$:
\begin{align}
\label{sampling}
p(\boldsymbol{x}_{s:s+H}) \propto \exp\left(\eta_s \hat{Q}(\boldsymbol{x}_{s:s+H})\right)
\end{align} 
At each time $t$, the decision is governed by the learner trained on observations up to $t-1$ (see Algorithm~\ref{s:algorithm_l4dc}). While we use an exhaustive search here, environments with larger state-action spaces can utilize more efficient search heuristics, such as Monte Carlo Tree Search (MCTS)~\citep{Bruegmann1993mcgo,Kocsis2006uct,Coulom2007mcts,Silver2016NatureGo,silver2017masteringchessshogiselfplay}.

\section{Experiments for Prospective Learning with Control}
\label{sec:exp_val}

To validate our framework, we devise a specific prospective learning problem called prospective foraging. 
Foraging is informally defined as the sequential search for resources~\citep{Barack2024-sx}. It is a central behavior of all agents, be they living organisms, AI bots, or robots. Standard formalizations of foraging assume that the environment is essentially stationary. For example,  without the single agent's intervention, resources stay available indefinitely. This is a poor model of real-world foraging because, typically, resources are depleted, rather than persistent. We therefore introduce prospective foraging as a special instance of a prospective learning with control problem. 

\subsection{Prospective foraging}
\label{s:prospective-foraging}

Informally, we assume an agent is navigating an environment, and resources (also called affordances~\citep{Gibson2014-or} or rewards~\citep{Sutton2018-gx}) appear and disappear in that environment at various locations.  At any given time, the agent may move its position, and the loss is the negative reward associated with the agent's current location.

\begin{SCfigure}
  \centering
  \caption{1-D foraging environment. An agent moves along a $1 \times 7$ linear track with two reward patches (A, B). Rewards alternate between the two patches over time, and the currently active patch’s reward decays exponentially.}
  \includegraphics[width=0.4\linewidth]{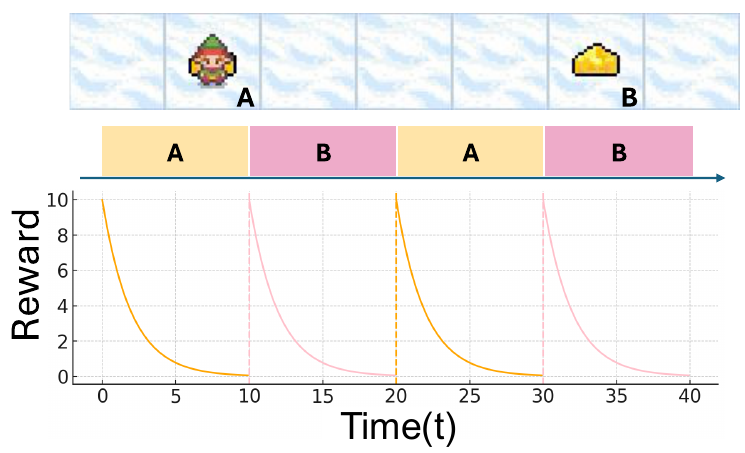}
  \label{f:foraging}
\end{SCfigure}

\paragraph{Data} Assume that $\XX$ is a $H \times W$ grid, and let  $H=1$, so the environment is a chain (see Figure~\ref{f:foraging}). Thus, $x_s \in \{1,\dots,W\}$, and $y_s \in \Real_+^W$.  We encode position via a one-hot encoding, thereby $x_s$ becomes a binary $W$-dimensional vector.   
We also embed time into a $50$-dimensional vector using the classical `positional' encoding strategy, as in~\citet{De-Silva2024-jj}.

\paragraph{Hypothesis} At each time step, the hypothesis can move the agent by one cell in the grid in either direction, assuming it is not at a boundary, or stay put.  At a boundary, it can only move nowhere or in the opposite direction. Formally, $h(x_s,s) = x_{s+1} \in \XX_s$, where $\XX_s$ could be either $ \{x_s-1, x_s, x_s+1\}$, $ \{x_s-1, x_s\}$, or $ \{x_s, x_s+1\}$.

\paragraph{Loss} The loss is defined as $-\sum_{k=t+1}^{\infty} \gamma^{k - t - 1} y_k$, with the temporal discounting factor of $\gamma$.

\paragraph{Assumptions} Let $F_{\bY}$ denote the prior distribution of $\bY$. Assume that $\bY$ is independent of $\bX$, i.e. $F_{\bY | \bX} = F_{\bY}$. Let $y_s \in \Real_+^W$ be a $ W$-dimensional non-negative vector, where each element encodes the reward that the agent would receive at location $x_s$ at time $s$. Assume that for two locations, $A$ and $B$, there are rewards sometimes, but for the other $W-2$ locations, there are never any rewards. 
For locations $A$ and $B$, assume that rewards arrive on a schedule. At every $r$ time steps, location $A$ receives a boost of one reward, which then decays exponentially down to zero at a rate of $\tau$. Location $B$ has the same temporal dynamics, but shifted by $r / 2$ time steps. Thus, 
 $y_s(A)=e^{-(s \mod r )/\tau}$, $y_s(B)=e^{-((s+r/2) \mod r )/\tau}$, 
 where $s \mod r$ indicates the modulus function.
The dynamics of the conditional distribution, $F_{\bX | \bY}$, are a function of the learned hypotheses, $\hat{h}$. 



We implement and evaluate the prospective learning controller (PLuC) within the 1-D foraging environment described in~\Cref{s:prospective-foraging} and~\citet{Bai2026-ik}: the agent forages along a $1 \times 7$ linear track containing only two reward patches, A and B, located three grid cells apart, as shown in~\Cref{f:foraging}. Every $r=10$ timestamps, rewards alternatively appear at A and B. The agent's goal is to maximize the total reward in its single lifetime, meaning there are no resets within each run.

We define the prospective regret of a hypothesis $h_t$ as the difference between the prospective risk returned by hypothesis $h_t$ and Bayes optimal risk: $\Delta{h_t} = R_t(h_t)-R_t^*$. In practice, one cannot evaluate regret over an infinite future. Therefore, we compute the prospective regret rate by evaluating it over a finite horizon of \(T\) time steps and taking the average:
$(R_{t,T}\!\big(h_t\big) - R_{t,T}^{\star}) / T$, where $R_{t,T}$ is the finite horizon prospective risk. In addition, since we only have one trajectory after time $t$, we replace the risk by the prospective loss, so $\Delta{h_t} \approx \frac{1}{T}\sum_{i=s}^{s+T} w_{i-s} (y_s(x^*_s) - y_s(x_s))$,
where $(x^*_s,x^*_{s+1},...,x^*_{s+T})$ are the states returned by the Bayes-optimal hypothesis at the corresponding times.

\subsection{Numerical Results}
\begin{figure}[!t]
    \centering
    \includegraphics[width=0.81\linewidth]{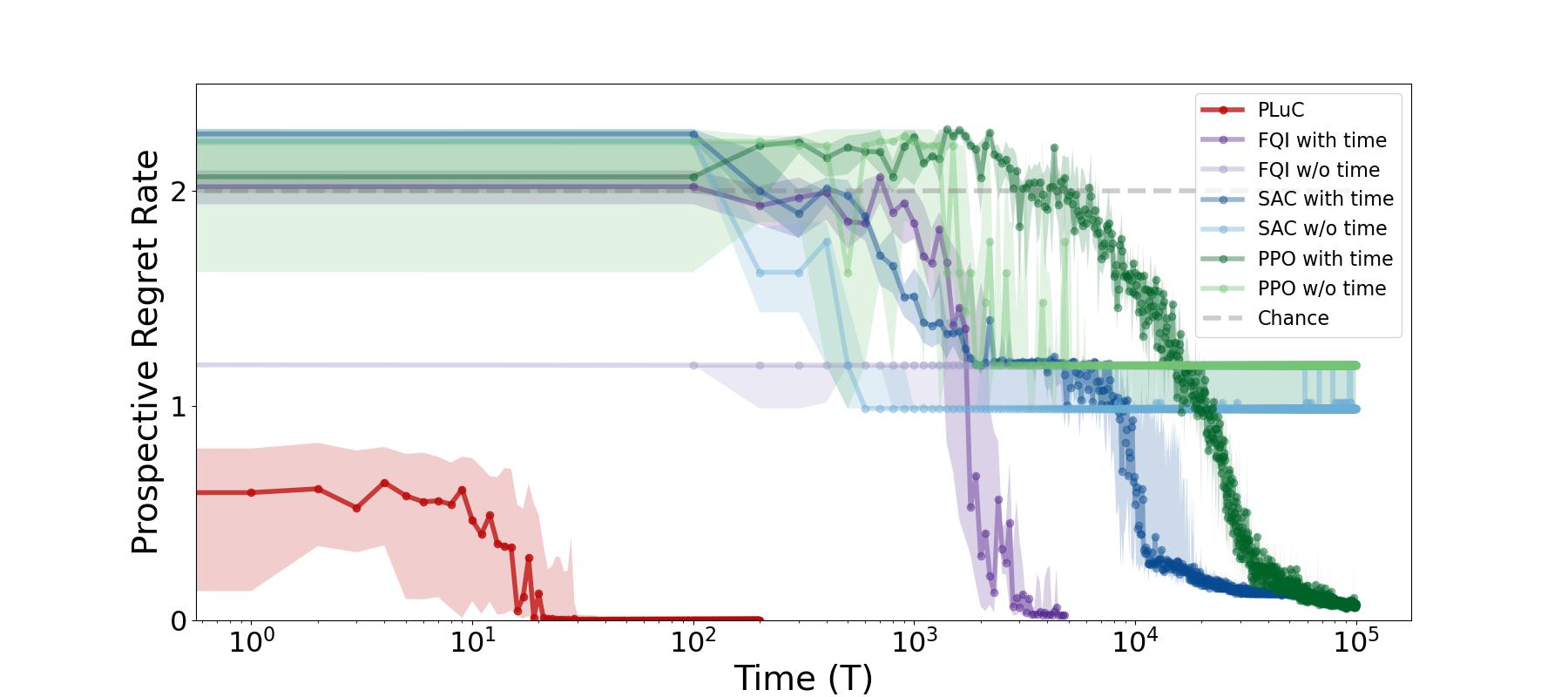}
    \caption{\textbf{PLuC converges to Bayes optimal regret more efficiently than RL baselines on the 1-D foraging task.} The prospective regret rate of PLuC (red), time-aware FQI (FQI with time, blue, our invention to improve FQI), time-agnostic FQI( FQI w/o time, light-blue), time-aware SAC (SAC with time, purple-red), time-agnostic SAC( SAC w/o time, lavender), time-aware PPO (PPO with time, dark-green), and time-agnostic PPO (PPO w/o time, light-green). While PLuC, time-aware FQI, time-aware SAC, and time-aware PPO converge to having zero regret, PLuC is orders of magnitude more efficient than any of them. And time-agnostic variants converge to a sub-optimal regret regardless of the time spent in interaction.}
    \label{f:proforg_fqi}
\end{figure}

After an initial warm-up phase, we let the PLuC learn in an online manner as it interacts with the environment over time (\Cref{f:proforg_fqi}). The procedure follows the steps detailed in~\Cref{s:algorithm_l4dc}. In the results presented here, however, we use a deterministic state-selection rule, where we choose the next state that yields the highest predicted prospective reward. Additional discussion of the Boltzmann sampling strategy described in the~\Cref{s:algorithm_l4dc} is provided in the~\Cref{sec:sampling}. During the online updating stage, we retrain regressors with all data collected so far for each update. We compute the normalized prospective regret of PLuC over time, and compare it to Fitted Q-Iteration (FQI)~\citep{ernst2005tree,munos2008finite}, Soft Actor-Critic (SAC)~\citep{haarnoja2018soft,haarnoja2018softa}, and Proximal Policy Optimization Algorithm (PPO)~\citep{schulman2017ppo}, three classical reinforcement learning algorithms, and the time-aware variants that we introduce here to adapt them to non-stationary environments (see~\Cref{sec:rl-baselines,alg:online-fqi,alg:online-sac,alg:online-ppo}). In both PLuC and FQI variants, we use random forests as the regressors. 

\begin{figure}[!t]
    \centering
    \includegraphics[width=0.69\linewidth]{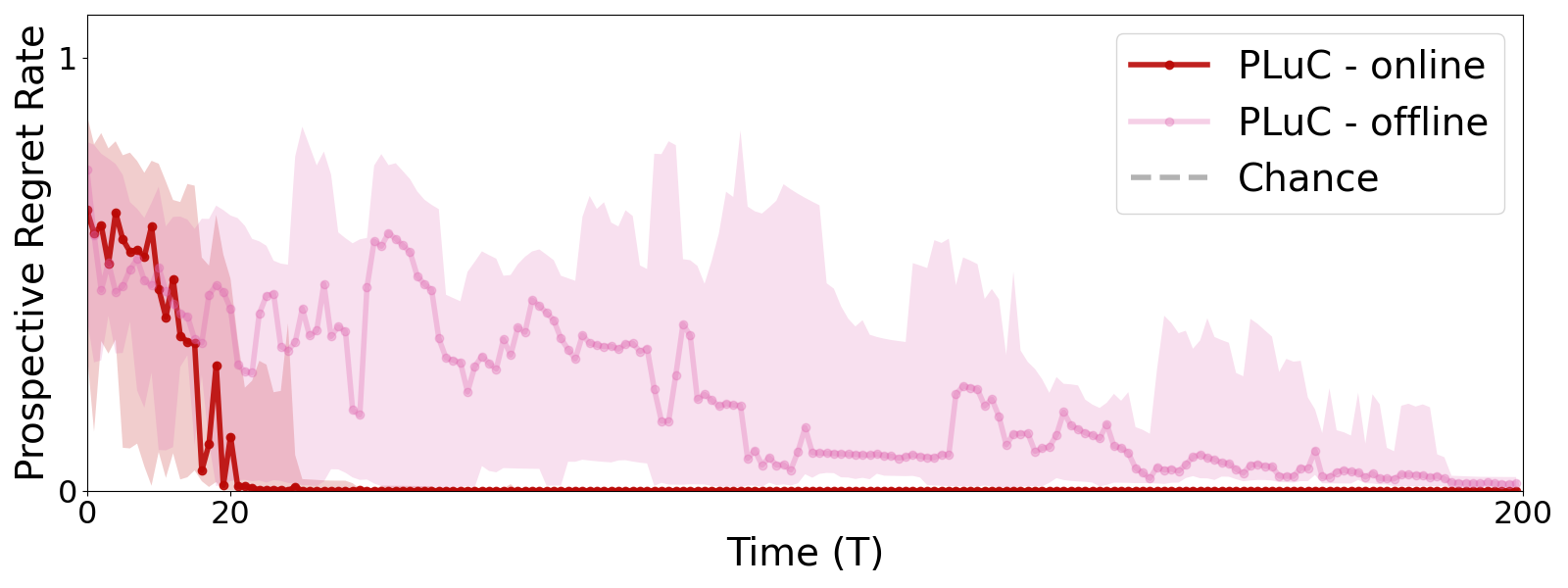}
    \caption{\textbf{PLuC online is several fold more efficient than offline.} Prospective regret rate for PLuC for online (red) and offline (pink). After warm-starting with 200 time steps, the online one converges in 20 time steps, whereas the offline one requires about 4$\times$ more data to converge. }
    \label{f:proforg_online}
\end{figure}

\paragraph{Online or Offline?} Building on the online formulation, we compare the online and offline PLuC, under the same environment settings and parameters for training and inference (\Cref{f:proforg_online}). The only difference is the next state exploration: online-PLuC chooses the next step using its currently learned policy, while offline-PLuC explores the next step uniformly at random. 

\begin{figure}[!t]
    \centering
    \includegraphics[width=0.69\linewidth]{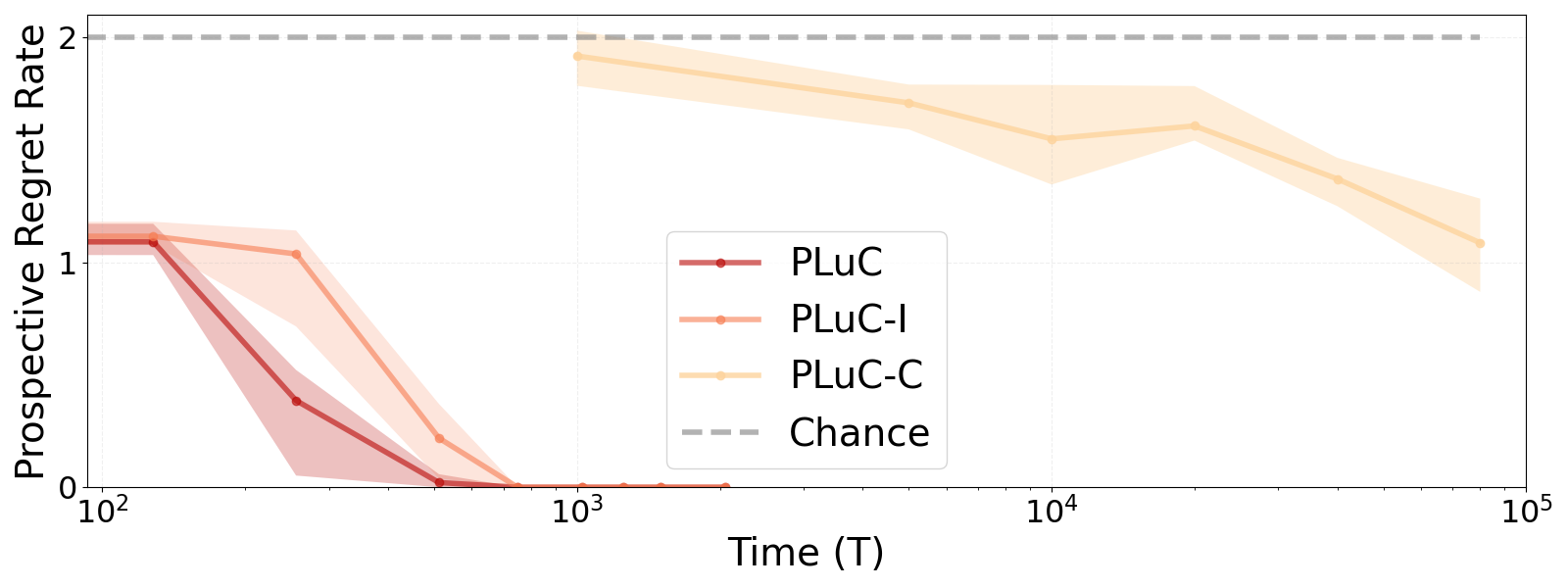}
    \caption{Prospective regret rate for PLuC (red), PLuC-I (orange), and PLuC-C (yellow). Removing either component reduces performance relative to PLuC.}
    \label{f:proforg_IC}
\end{figure}

\paragraph{Instantaneous, Cumulative or Both?} PLuC uses two learners during training and inference phases, $g^i$ for instantaneous loss, and $g^c$ for future cumulative loss. During inference, we combine a finite-horizon ``lookahead'' from $g^i$ with a terminal cost approximation from $g^c$. We compare offline invariants of Instantaneous-only (PLuC-I), Future-only (PLuC-C), and combined (PLuC) (\Cref{f:proforg_IC}). Using the terminal cost alone (PLuC-C) converges inefficiently, but we find adding it to the instantaneous loss accelerates learning versus instantaneous-only.

\begin{figure}[!t]
    \centering\includegraphics[width=0.69\linewidth]{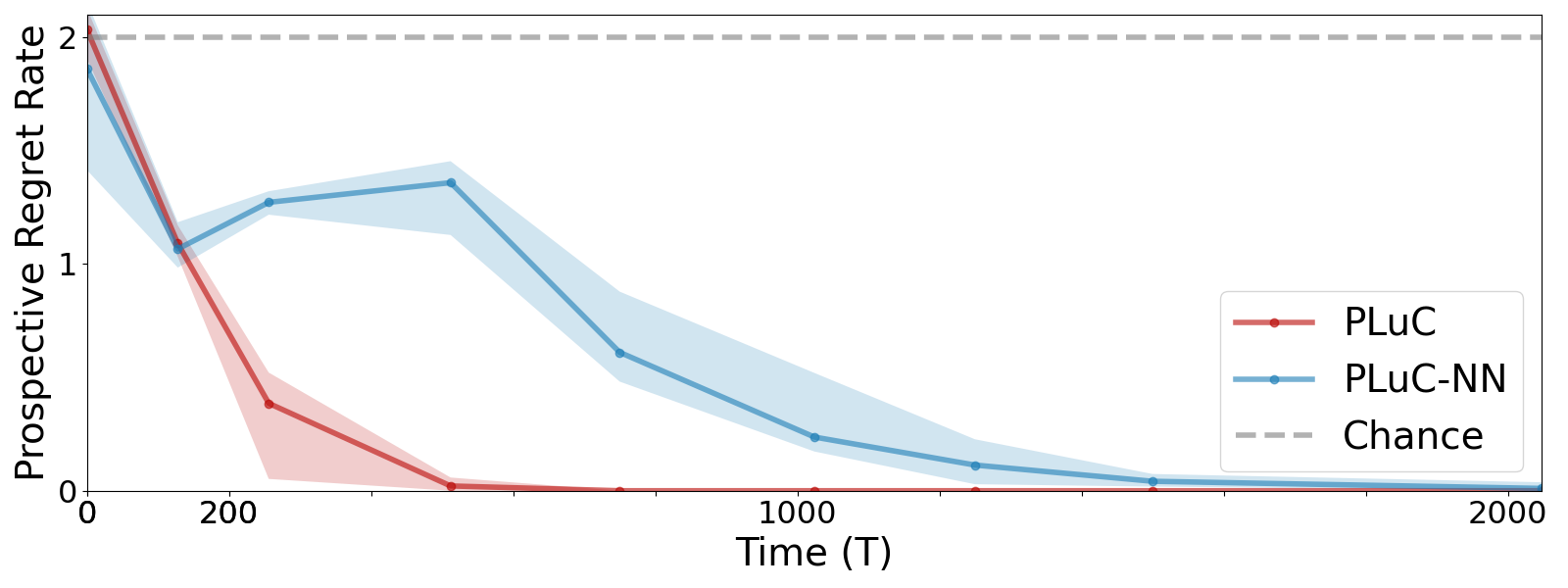}\caption{\textbf{PLuC with decision forests is 4x more efficient than with neural networks.} Prospective regret rate for PLuC with Gradient-Boosted Trees (red) and MLP Regressor (blue). While PLuC is 4x more efficient,   PLuC-NN does converge as well. }
\label{f:proforg_dfnn}
\end{figure}

\paragraph{PLuC with Neural Network} In~\Cref{f:proforg_dfnn}, we show that neural networks can also be used as the regressor in PLuC. While the PLuC with a neural network (specifically, a multi-layer perceptron with two hidden layers comprising 128 units) converges eventually, the convergence is about an order of magnitude slower than the PLuC employing a random forest as the regressor.


\section{Discussion of Prospective Learning with Control}
In this work, we introduce the framework of prospective learning with control and evaluate it on a foraging task, a simple but illustrative non-stationary control problem. Our results show that our algorithm, PLuC, can learn to plan ahead optimally. We also observe that standard RL baselines, when applied without modification, fail to converge to the optimal policy despite the simple nature of the one-dimensional foraging environment, which is an expected outcome given their stationary assumptions. Time-aware modifications improve convergence but at a significant sample-efficiency cost relative to PLuC. Much work remains to be done, as our experiments are currently restricted to a simple environment with deterministic dynamics and periodic reward. We believe prospective learning with control is an initial step towards understanding and utilizing optimal control of the future in both natural and artificial intelligence. We leave generalizing to continuous action, state, and time spaces, and deploying our algorithms with transition dynamics in complex real-world environments, as future work.

\clearpage
\bibliography{backmatter/bib}

\clearpage
\appendix
\section{Prospective Learning without control (PLiC)}
\label{sec:dynamic}

Here we briefly review the prior work on this topic, which is called "prospective learning"~\citep{De_Silva2023-al, De-Silva2024-jj, Bai2026-ik} (PL), modifying notation slightly for convenience.  In retrospect, it would more specifically be called prospective learning without control, or \textbf{P}rospective \textbf{L}earning m\textbf{i}nus \textbf{C}ontrol (\textbf{PLiC}).

\paragraph{Data}
Let the input and output be denoted by $x_t \in \XX$ and $y_t \in \YY$, respectively. Let
$z_t = (x_t, y_t)$. We will denote the observed data, $z_{\leq t} = \{z_1, \dots, z_t\}$, and the unobserved data, $z_{>t}$. In contrast to PAC learning, $t$  is not just a dummy variable, but rather, indexes time.  Let $t \in \TT \subseteq \naturals$, where $\naturals=\{1,2,\dots\}$ are the counting numbers, and let $|\mc{T}| = T \leq \infty$ be the total number of time steps. 
We therefore define the data triple $(x_t,y_t,t)$.

\paragraph{Hypothesis} The hypothesis space generalizes the notion from PAC learning. Here,  $h_t : \XX \times \TT \to \YY$ is a hypothesis in $\mc{H}_t \subseteq  \{h : \XX \times \TT \to \YY \}$, so $h_t$ infers an output $y_{t}$ for a given input $x_t$ at time $t$.%
 Denote the hypothesis sequence,   $h \equiv (h_1,h_2,\dots)$, where each element of this sequence $h_t: \XX \mapsto \YY$.
At times, we will consider a sequence of nested hypothesis spaces, $\HH_1 \subseteq \HH_2 \subseteq \dots \subseteq \HH$.

\paragraph{Loss}
The instantaneous loss, $\bar{\ell}: \HH \times \ZZ \to \Reals$, is a map from the hypothesis and current data to the reals.  In an abuse of notation, we write  $\bar{\ell}(h_t,z_t) \equiv \bar{\ell}( h(x_t,t),y_t)$.
In all real-world problems, we care about the integrated future loss. 
Let $w_t$ be a non-increasing non-negative weighting function that sums to one, that is $\sum_t w_t = 1$ and $0 \leq w_t \leq 1\, \forall t$.  We thus 
define loss at time $t$ as $\ell_t(h, z_{>t}) = \sum_{s>t} w_{s-t} \bar{\ell}(h(x_s,s),y_s)$, which is the weighted cumulative loss over all the future data.%
\footnote{If $T$ is infinite, we replace the sums with limiting sums, e.g.,  $\lim_{T\to \infty} \sum_{t=1}^T w_t$, to ensure that the sum converges.}$^,$%
\footnote{Note that if we let $w_t$ be a constant function of $t$,  we recover something equivalent to the loss in PAC learning.
}
To fully specify loss, therefore, requires specifying a temporal discounting function, $\mb{w} \equiv (w_1, w_2, \dots)$.  For example, we could let $\mb{w}$ be an exponentially decaying function, $w_{s-t}= \gamma^{s-t}$, where $\gamma \leq 1$.\footnote{This is a slight variation of how prospective loss was previously defined in~\citet{De-Silva2024-jj}.} 

\paragraph{Learner}
Let a learner, $\mc{L}$ maps from the data to a hypothesis, $\mc{L}(\mc{D}_t) = \hat{h}^t$, where $\mc{D}_t = (x_s,y_s,s)_{s\leq t}$, and $\hat{h}^t$ is the estimated hypothesis sequence obtained by the learner at time $t$.%
\footnote{Formally, we assume $h$ is a random variable and $h\in \sigma(Z_{\leq t})$ where $\s(\cdot)$ denotes the filtration (an increasing sequence of sigma-algebras) of the stochastic process $Z$}.

\paragraph{Assumptions}
Let $\bX=X_{0<t}$ and $\bY=Y_{0<t}$ be stochastic processes, taking values $\bX=x_{0<t}$ and $\bY=y_{0<t}$.  
%
Assume $(\bX,\bY) \sim F \in \mc{F}$.  In general, we place no assumptions on $\mc{F}$.

\paragraph{Theoretical Goal} 
Define risk as expected loss, $R_t(h) = \E_F[\ell_t(h, Z) \mid z_{\leq t}]  = \int \ell_t (h, Z ) \ \mathrm{d} \mathbb{P}_{Z \mid z_{\leq t}}$.\footnote{This is a slight abuse of notation, because previously $\ell$ mapped from a hypothesis and data corpus, but now we are saying that it maps from a hypothesis and a collection of random variables. 
We have used the shorthand  $\E[Y \mid x]$ for  $\E[Y \mid X = x]$. 
Note that $\mathbb{E}[Z | z_{\leq t}]$ is equivalent to $\mathbb{E}[Z_{>t} | z_{\leq t}]$ because the past $z$'s are given. 
} 
%
Let $R_t^* = \min_{h \in \mc{H}_t} R_t(h)$ be the (Bayes) optimal risk.\footnote{In a slight abuse of notation, we use $\min$ in place of $\inf$ even if the $\min$ might not exist. Here we use the shorthand  $\mc{H}_t \equiv \sigma(Z_{\leq t})$.}
 %
 
We seek  a learner, $\mc{L}$ with the following property: for any $F \in \mc{F}$, and $\eps,\delta > 0$, 
there exists a   $t'(\eps,\delta)$ such that the learner outputs a hypothesis $\hat{h}^t$ satisfying $\P[ R_t (\hat{h}^t) - R_t^* < \eps] \geq  1- \delta$, for any $t > t'(\eps,\delta)$. 

One implication of the above desiderata is that the estimated hypothesis, $\hat{h}^t$, is a consistent estimator of the risk~\citep{Shalev-Shwartz2014-zo}.  In other words,  $R(\hat{h}^t) \to R^*_t$ as $t \to \infty$.  
The above goal, however, is not an explicit objective function.  In practice, therefore, we specify an objective function to minimize.  The hope is that if we minimize this objective, then eventually, the resulting estimand satisfies the above desiderata. We seek to minimize  the weighted sum of  instantaneous losses on  \textit{future} data: 
\begin{equation} \label{eq:perm}
\hat{h}^t =  \argmin_{h \in \mc{H}_t} R_t(h)  
= \argmin_{h \in \mc{H}_t} \int \ell_t (h, Z ) \ \mathrm{d} \mathbb{P}_{Z \mid z_{\leq t}} 
= \argmin_{h \in \mc{H}_t} \int \sum_{s>t} w_{s-t} \bar{\ell}(h(x_s,s),y_s) \ \mathrm{d} \mathbb{P}_{Z \mid z_{\leq t}}. 
\end{equation}

If $z$ is a discrete set, the above integral becomes a sum.  Let $p_z$ denote the probability of a particular sequence $z \in \ZZ$.  With this notation,~\Cref{eq:perm} becomes
\begin{equation} \label{eq:dperm}
\hat{h}^t  
= \argmin_{h \in \mc{H}_t} \sum_{z \in \ZZ : z_{\leq t}} p_z  \ell_t (h, Z )  
= \argmin_{h \in \mc{H}_t} \sum_{z \in \ZZ : z_{\leq t}} p_z \sum_{s>t} w_{s-t} \bar{\ell}(h(x_s,s),y_s). 
\end{equation}

\paragraph{Empirical Goal}
In practice, in general, we cannot minimize~\Cref{eq:perm} because it requires solving an intractable integral, and we typically cannot solve~\Cref{eq:dperm}, because it requires sampling too many possible futures (there are exponentially many even if $z$ is binary).  Moreover, we cannot minimize~\Cref{eq:perm,eq:dperm} because they both require sampling from an unknown distribution, $\mathbb{P}_{Z \mid z_{\leq t}}$.  To address the first issue, we use a Monte Carlo approximate sum, sampling only a subset of possible futures. To address the second issue, rather than sampling from the unobserved future, we sample only from the observed past.  With these two approximations in hand, we have  
\begin{equation}    \label{eq:perm1}
\hat{h}^t   \approx 
\argmin_{h^t \in \mc{H}_t}  \sum_{t'>0}^t \ell_{t'}(h,z_{>t'})
\approx \argmin_{h^t \in \mc{H}_t} \sum_{t' > 0 }^t \sum_{s> t'}^t w_{s-t'} \bar{\ell}(h^{t'}(x_s,s),y_s).    
\end{equation}
\textbf{Main result} The main result of~\citet{De-Silva2024-jj} is approximating a variant of~\Cref{eq:perm},  under suitable assumptions, satisfies our theoretical goal.  The key assumptions are (a) consistency and (b) uniform concentration.  The focus in that work was on settings in which our decisions $\hat{h}^t(x_s,s)$ had no impact on future outcomes or distributions.  In such cases, we can ignore the sum in~\Cref{eq:perm},
because all that matters for the current decision is the instantaneous loss.  Here, we are interested in general settings where the decisions also impact future distributions and losses. 


\paragraph{Relationship to prospective learning without control}
%
The observed \textit{data} triples in PLuC are $(x_s,y_{s+1}(x_{s+1}),s)$, rather than $(x_s,y_s,s)$ in PLiC. Note that these data triples differ in two ways.  First, the index on $y$ changes from $s$ to $s+1$ as we go from PLiC to PLuC; this is required because the rewards are necessarily a function of the action of the agent.  Second, rather than the entire vector of output being observed as in PLiC, in PLuC,  we only obtain a partial observation: the reward at the location we happen to currently inhabit.  This renders PLuC  closely related to causal inference, in which the outcomes are defined as a function of the intervention (or treatment), and therefore only partially observed, leading to the potential outcomes framework~\cite{Neyman1923-aw, Rubin1974-gt}. 
\textit{Hypotheses} are no longer maps from $\XX \times \TT$ to $\YY$, rather, they map to $\XX$ and $\Real_+$.
\textit{Loss} in PL with or without control is the cumulative instantaneous loss. However, in PLiC, instantaneous loss is typically an error signal, e.g., the difference between the predicted output and the actual output.  In PLuC, loss is just a negative reward.  Moreover, loss is not observed for any of the decisions that the agent did not make, necessitating the use of some surrogate loss function in practice.  
\textit{Learners} map from data to hypotheses, though the data are subtly different due to the time indices and partial observation.
\textit{Assumptions} can be the same, though note that the future data are now necessarily a function of the learner and hypothesis in PLuC, whereas they were not necessarily in PLiC. 
The \textit{theoretical goal} is the same, modulo replacing the loss from PLiC with that from PLuC.  The \textit{empirical goal} changes somewhat similarly,  insofar as in PLuC we cannot directly solve~\Cref{eq:PFgoal}, so we use a surrogate optimization problem,~\Cref{eq:perm2}.



\section{Theoretical Results}
\label{s:proofs}
\begin{lemma}
    Suppose $\lbrace Z_t \rbrace_{t=1}^{\infty}$ is a stochastic process and let $\lbrace \mathcal{H}_t \rbrace_{t=1}^{\infty}$ be an increasing hypothesis class, such that there exists $h_t \in \mathcal{H}_t$ which satisfies
    \begin{equation}
        \lim_{t \to \infty}
        \mathbb{E}[R_t(h_t)-R_t^*]=0.
    \end{equation}
    Additionally, suppose $\lbrace u_t \rbrace_{t=1}^{\infty}$ is  a sequence such that $u_t \to \infty$ as $t \to \infty$, and $u_t \leq t$ for all $t$. Define
    \begin{equation*}
        e_m(h)=
        \sum_{s=1}^m w_{s} \tilde{y}_{s+1}(x_{s+1}).
    \end{equation*}
    
    Then, there exists $\tilde{h}_{t} \in \mathcal{H}_t$ such that
    \begin{equation*}
    \lim_{t \to \infty}
        \mathbb{E}
        \left[
\sup_{u_t \leq m \leq \infty}
e_m(\tilde{h}_{t})|Z_{\leq t}
        \right]-R^*_t =0.
    \end{equation*}
\end{lemma}
\textbf{Proof.} There exists $h_t \in \mathcal{H}_t$ such that 
\begin{equation*}
    \lim_{t \to \infty} \mathbb{E}[R_t(h_t)-R_t^*]=0.
\end{equation*}
Note that $R_t(h) \geq R_t^*$ almost surely. Choose a subsequence $\lbrace
j_k \rbrace_{k=1}^{\infty}$ such that 
\begin{equation*}
    \mathbb{E}[R_{j_k}(h_{j_k})-R_t^*] \leq \frac{1}{4^k}. 
\end{equation*}
Now,
\begin{equation*}
\begin{aligned}
    \mathbb{E}[R_t(h_t)-R_t^*] &= 
    \mathbb{E}
    \left[
    \mathbb{E}[\lim \sup_{m \to \infty} e_m(h_t)| Z_{\leq t}]-R_t^*
    \right] \\
    &= \mathbb{E}\left[
    \mathbb{E}[\lim_{i \to \infty} \sup_{u_i \leq m \leq \infty} e_m(h_t)|Z_{\leq t}]
    -R_t^*
    \right] \\
    &= \lim_{i \to \infty}
    \mathbb{E}
    \left[
    \mathbb{E}[
    \sup_{u_i \leq m \leq \infty} e_m(h_t)|Z_{\leq t}
    ]-R_t^*
    \right]
\end{aligned}
\end{equation*}
Thus, for every $k$, there exists integer $i_k$ such that 
\begin{equation*}
    \mathbb{E}
    \left[
    \mathbb{E}[
    \sup_{u_{i_k} \leq m \leq \infty} e_m(h_{j_k})|Z_{\leq j_k}
    ]-R_{j_k}^*
    \right] \leq \mathbb{E}[R_{j_k}(h_{j_k})-R_{j_k}^*]+\frac{1}{4^k} \leq \frac{2}{4^k}.
\end{equation*}
Note that,
\begin{equation*}
\begin{aligned}
    \mathbb{E}
    \left[
    \sup_{u_i \leq m \leq \infty} e_m(h_t)|Z_{\leq t}
    \right] \geq 
    \mathbb{E}
    \left[
    \sum_{s=t+1}^m w_{s-t}\tilde{y}_{s+1}(x_{s+1})|Z_{\leq t}
    \right] \geq R_t^*.
\end{aligned}
\end{equation*}
Using Markov's Inequality,
\begin{equation*}
\begin{aligned}
    &\sum_{k=0}^{\infty}
    \mathbb{P}
    \left(
\mathbb{E}
\left[
\sup_{u_{i_k} \leq m \leq \infty} e_m(h_{j_k})|Z_{\leq j_k}
\right] - 
R_{j_k}^*
>2^{\frac{1}{2}-k}
    \right) \\
    &\leq 
    \sum_{k=0}^{\infty}
    \frac{1}{2^{\frac{1}{2}-k}}
    \mathbb{E}
    \left[
\mathbb{E}
\left[
\sup_{u_{i_k} \leq m \leq \infty} e_m(h_{j_k})|Z_{\leq j_k}
\right]-
R_{j_k}^*
    \right] \\
    &\leq \sum_{k=0}^{\infty}
    \frac{1}{2^{\frac{1}{2}-k}}
    \frac{2}{4^k} \\
    &< \infty. 
\end{aligned}
\end{equation*}
Thus, by Borel-Cantelli Lemma, there exists an integer $k_0$ such that for all $k \geq k_0$, 
\begin{equation*}
    \mathbb{E}
    \left[
\sup_{u_{i_k} \leq m \leq \infty} e_m(h_{j_k})|Z_{\leq j_k} 
    \right] - R_{j_k}^*
    \leq 
    2^{\frac{1}{2}-k}
    . 
\end{equation*}
Now, we define
$k_t=\max \lbrace
k \in \mathbb{N} \cup 
\lbrace 0 \rbrace :
\max \lbrace
i_k,j_k
\rbrace \leq t
\rbrace$.
Note that $k_t \to \infty$ as $t \to \infty$, because $i_k,j_k \leq \infty$. Define $\alpha_t=2^{\frac{1}{2}-k_t}$, and observe that $\lim_{t \to \infty} \alpha_t=0$. Let $t_0 \in \mathbb{N}$ be a random variable such that for all $t \geq t_0$, $k_t \geq k_0$, and thus,
\begin{equation*}
    \mathbb{E}
    \left[
    \sup_{
u_{i_{k_t}} \leq m \leq \infty
}
e_m(h_{j_{k_t}})
\big| Z_{\leq j_{k_t}}
    \right]
    -R^*_{j_{k_t}}
    \leq \alpha_t.
\end{equation*}
Choose $\tilde{h}_{t}=h_{j_{k_t}}$ for all $t$. Note that $j_{k_t} \leq t \implies \mathcal{H}_{j_{k_t}} \subset \mathcal{H}_t$. Hence, for $t \geq t_0$,
\begin{equation*}
    \mathbb{E}
    \left[
\sup_{u_t \leq m \leq \infty}
e_m(\tilde{h}_{t})|Z_{\leq j_{k_t}}
    \right]-R_{j_{k_t}}^*
    \leq 
    \mathbb{E}
    \left[
\sup_{u_{i_{k_t}} \leq m \leq \infty}
e_m(\tilde{h}_{t})|Z_{\leq j_{k_t}}
    \right]-
R^*_{j_{k_t}}
    \leq \alpha_t.
\end{equation*}
Since $\alpha_t \to 0$ as $t \to \infty$, we have,
\begin{equation*}
    \lim_{t \to \infty}
    \bigg(
    \mathbb{E}
    \left[
\sup_{u_t \leq m \leq \infty}
e_m(\tilde{h}_{t})|Z_{\leq j_{k_t}}
    \right]-R^*_{j_{k_t}}
    \bigg) =0.
\end{equation*}

\begin{theorem}
    Consider a family of stochastic processes $(\mathbf{X}_t,\mathbf{Y}_t)_{t>0}$. Suppose there exists a hypothesis class such that $\mathcal{H}_1 \subseteq \mathcal{H}_2 \dots  $ such that 
    \begin{equation*}
        \lim_{t \to \infty}
        \left[
\inf_{h \in \mathcal{H}_t} R_t(h)-R^*_t
        \right]=0. 
    \end{equation*}
Also, there exist sequences $\lbrace u_t \rbrace_{t=1}^{\infty}$ and $\lbrace \xi_t \rbrace_{t=1}^{\infty}$ satisfying $\lim_{t \to \infty} \xi_t=0$, and $u_t \leq t$, such that
\begin{equation*}
    \mathbb{E}
    \left[
\max_{h \in \mathcal{H}_t}
\left|
\sum_{s=t+1}^{\infty}
w_{s-t} \tilde{y}_{s+1}(x_{s+1}) -
\max_{u_t \leq m \leq t}
\sum_{s=1}^{m} w_{s} \tilde{y}_{s+1}(x_{s+1})
\right|
    \right] \leq \xi_t.
\end{equation*}
Then, there exists a sequence $\lbrace i_t \rbrace_{t=1}^{\infty}$ such that 
\begin{equation*}
    \hat{h}^{(t)}=
    \argmin_{h \in \mathcal{H}_{i_t}} \max_{u_{i_t} \leq m \leq t}
    \sum_{s=1}^m 
    w_{s} \tilde{y}_{s+1}(x_{s+1})
\end{equation*}
reaches the Bayes' optimal risk.
\end{theorem}
\textbf{Proof.} 
We consider a subsequence $\lbrace i_t \rbrace_{t=1}^{\infty}$ such that 
$\xi_{i_t} \to 0$ exponentially so that $\sum_{t=1}^{\infty} \xi_{i_t} < \infty$.
From Markov's Inequality,
\begin{equation*}
\begin{aligned}
&\sum_{t=0}^{\infty}
    \mathbb{P}
    \left[
\max_{h \in \mathcal{H}_{i_t}}
\left|
\Bar{l}_t(h,Z)-
\max_{u_{i_t} \leq m \leq i_t}
e_m(h)
\right|
> \sqrt{\xi_{i_t}}
    \right]  \\
    &\leq 
    \sum_{t=0}^{\infty}
    \frac{1}{\sqrt{\xi_{i_t}}}
    \mathbb{E}
    \left[
  \max_{h \in \mathcal{H}_{i_t}}
\left|
\Bar{l}_t(h,Z)-
\max_{u_{i_t} \leq m \leq i_t}
e_m(h)
\right|  
    \right] \\
    &\leq 
    \sum_{t=0}^{\infty}
    \sqrt{\xi_{i_t}}
    \\
    &< \infty.
\end{aligned}
\end{equation*}
By the Borel-Cantelli Lemma, 
there exists $t_1 \in \mathbb{N}$, such that for all $t \geq t_1$,
\begin{equation*}
    \max_{h \in \mathcal{H}_{i_t}}
    \left|
\Bar{l}_t(h,Z)-
\max_{u_{i_t} \leq m \leq i_t}
e_m(h)
    \right|
    \leq \sqrt{\xi_{i_t}}.
\end{equation*}
Let $j_t=\max \lbrace
t': i_{t'} \leq t
\rbrace$. Observe that $i_{j_t} \to \infty$ as $t \to \infty$. Since $i_{j_t} \leq t$,
\begin{equation*}
    \Bar{l}_t(h,Z)-
    \max_{u_{i_{j_t}} \leq m \leq t}
    e_m(h)
    \leq \Bar{l}_t(h,Z)-\max_{u_{i_{j_t}} \leq m \leq i_{j_t}} e_m(h).
\end{equation*}
Construct a random variable $t_2$ such that for all $t \geq t_2$,$j_t \geq t_1$. 
Thus, for all $t \geq t_2$,
\begin{equation*}
\begin{aligned}
    &\max_{h \in \mathcal{H}_{i_{j_t}}}
    \left\lbrace 
\Bar{l}_t(h,Z)-
\max_{u_{i_{j_t}} \leq m \leq t} e_m(h)
    \right\rbrace \\
    &\leq 
    \max_{h \in \mathcal{H}_{i_{j_t}}}
    \left|
\Bar{l}_t(h,Z)-
\max_{u_{i_{j_t}} \leq m \leq i_{j_t}} e_m(h)
    \right| \\
    &\leq \sqrt{\xi_{i_{j_t}}}. 
\end{aligned}
\end{equation*}
Choose $i_t=i_{j_t}$ and notice that since $i_t \to \infty$, we have $i_t \leq t$. With probability one, for all $t \geq t_2$,
\begin{equation*}
    \max_{h \in \mathcal{H}_{i_t}}
    \left\lbrace 
\Bar{l}_t(h,Z)-\max_{u_{i_t} \leq m \leq t}
e_m(h)
    \right\rbrace
    \leq \sqrt{\xi_{i_t}}.
\end{equation*}
For $t \geq t_2$, 
\begin{equation*}
\begin{aligned}
    \Bar{l}_t(\hat{h}^{(t)},Z) &\leq 
    \max_{u_{i_t} \leq m \leq t} e_m(\hat{h}^{(t)})
    + \sqrt{\xi_{i_t}}
    \\
    &\leq \max_{u_{i_t} \leq m \leq t} e_m(h_t) + 
    \sqrt{\xi_{i_t}}
    \\
    &\leq \sup_{u_{i_t} \leq m \leq t}
    e_m(h_t) + 
    \sqrt{\xi_{i_t}}
\end{aligned}
\end{equation*}
Hence, 
\begin{equation*}
    \mathbb{E}[\Bar{l}_t(\hat{h}^{(t)},Z)|Z_{\leq t}] -
    \mathbb{E}
    \left[
\sup_{u_{i_t} \leq m \leq t} e_m(h_t)|Z_{\leq t}
    \right] \to 0
\end{equation*}
almost surely. By Lemma 3, we have, 
\begin{equation*}
    \mathbb{E}
    \left[
\Bar{l}_t(\hat{h}^{(t)},Z)|Z_{\leq t}
    \right] - R_t^*
    \to 0.
\end{equation*}
Hence, by the bounded convergence theorem, 
\begin{equation*}
    0= 
    \mathbb{E}
    \left[
\lim_{t \to \infty}
\left(
\mathbb{E}
\left[
\Bar{l}_t(\hat{h}^{(t)},Z)|Z_{\leq t}
\right]-R_t^*
\right)
    \right]=
    \lim_{t \to \infty}
    \mathbb{E}
    \left[
\mathbb{E}
\left[
\Bar{l}(\hat{h}^{(t)},Z)|Z_{\leq t}
\right]-R_t^*
    \right]. 
\end{equation*}

\section{Reinforcement Learning Baselines}
\label{sec:rl-baselines}

Let $s_t \in \mathcal{S}, a_t \in \mathcal{A}$ and $r_t \in \Reals$ denote the state, action, and reward at time $t$, respectively. 

\subsection{Fitted Q-Iteration}
\label{ssec:fqi}

Suppose we are given a dataset of interactions $\mc{D} = \{(s_t, a_t, r_t)\}_{t=1}^T$ collected using some stochastic behavior policy by interacting with a Markov Decision Process (MDP). The generic Fitted Q-iteration (FQI)~\citep{ernst2005tree,munos2008finite} algorithm approximates the state-action value function $Q: \mc{S} \times \mc{A} \to \Reals$ by running the following iterative condition. At iteration $k$, it constructs training targets
\[
y_t^{(k)} = r_t + \gamma \max_{a' \in \mathcal{A}} Q_k(s_{t+1},a'),
\]
from the dataset $\mc{D}$, and then it updates $Q_{k+1}$ by fitting a regressor from a model class $\mc{F}$ onto these targets.
\[
Q_{k+1} = \argmin_{f \in \mc{F}} \sum_{t=1}^T \left( f(s_t, a_t) - y_t^{(k)} \right)^2
\]
Once we have the approximated $\hat Q$-function, we use the greedy policy given by, 
\[
\pi(s) = \argmax_{a \in \mc{A}} \hat{Q}(s, a)
\]
In the time-aware variant of FQI, we make the $Q$-function a function of time $t$ in addition to the state $s$ and action $a$. We do this by slightly modifying the iterative condition as follows. At iteration $k$, it constructs training targets
\[
y_t^{(k)} = r_t + \gamma \max_{a' \in \mathcal{A}} Q_k(s_{t+1},a', t),
\]
from the dataset $\mc{D}$, and then it updates $Q_{k+1}$ by fitting a regressor from a model class $\mc{F}$ onto these targets.
\[
Q_{k+1} = \argmin_{f \in \mc{F}} \sum_{t=1}^T \left( f(s_t, a_t, t) - y_t^{(k)} \right)^2
\]
As before, once we have the approximated $\hat Q$-function, we construct a greedy policy by, 
\[
\pi(s, t) = \argmax_{a \in \mc{A}} \hat{Q}(s, a, t)
\]
However, unlike the static policy from generic FQI, the time-aware variant leads to a dynamic policy that varies with time. In our experiments, we use a Random Forest with $1000$ trees as the regressor of the FQI. We illustrate the pseudocode of online FQI in~\cref{alg:online-fqi}.

\begin{algorithm}[H]
\caption{Online Fitted Q-Iteration with $\epsilon$-greedy exploration}
\label{alg:online-fqi}
\begin{algorithmic}[1]
\State \textbf{Initialize} Foraging environment \texttt{env}, experience buffer $\mathcal{D}$, warmup size $W$, update interval $I$, exploration parameter $\epsilon$, action space $\mathcal{A} = \{-1, 0, 1\}$
\State $\hat Q \gets \text{None}$
\State \textbf{for} $t = 1$ to $T$
    \State  \qquad $s_t \gets \texttt{env}.\text{get\_current\_state()}$
    
    \State  \qquad Sample $u \sim \text{Uniform}[0,1]$
    \State \qquad \textbf{if} $u < \max(0.01, \epsilon \cdot 0.999^t)$ \textbf{or} $Q$ is None
        \State \qquad \qquad $a_t \gets \text{RandomChoice}(\mathcal{A})$ \Comment{Explore}
    \State \qquad \textbf{else}
        \State \qquad \qquad$a_t \gets \argmax_{a' \in \mathcal{A}} \hat Q (s, a', t)$ \Comment{Exploit}
    \State \qquad \textbf{end if}
    
    \State \qquad $(s_{t+1}, r_t) \gets \texttt{env}.\text{step}(a_t)$
    \State \qquad Store $(s_t, a_t, r_t, t)$ in $\mathcal{D}$
    
    \State \qquad \textbf{if} {$|\mathcal{D}| \geq W$ \textbf{ and } $(t+1) \bmod I = 0$}
        \State \qquad \qquad $\hat Q = \texttt{FQI}.\text{fit}(\mathcal{D})$
    \State \qquad \textbf{end if}
\State \textbf{end for}
\end{algorithmic}
\end{algorithm}

\subsection{Soft Actor-Critic}
\label{ssec:sac}

Soft Actor-Critic (SAC)~\citep{haarnoja2018soft, haarnoja2018softa} is an off-policy, model-free algorithm developed to perform entropy-regularized reinforcement learning, where the agent aims to maximize both cumulative return and the entropy of the policy;
\[
J(\pi) = \mathbb{E} \left[ \sum_{t=0}^\infty \gamma^t \left( r_t + \alpha H\left(\pi(\cdot \mid s_t\right) \right) \right]
\]
Here, $\alpha$ is the temperature parameter, $H(\cdot)$ is the entropy, and $\gamma$ is the discount factor. Maximizing the entropy of the policy alongside the cumulative reward encourages exploration and prevents premature convergence to suboptimal policies.~\Cref {alg:online-sac} illustrates the online-SAC algorithm used in this work. We build the time-aware SAC by making the policy, critics, and target critics functions of time $t$.

\begin{algorithm}[h]
\caption{Online Soft Actor-Critic}
\label{alg:online-sac}
\begin{algorithmic}[1]
\State \textbf{Initialize} Foraging environment \texttt{env}, experience buffer $\mathcal{D}$, Policy network $\pi_\theta$, Q-function/critic networks $Q_{\phi_1}$ and $Q_{\phi_2}$, batch size $B$
\State Set target critic parameters equal to main critic parameters $\phi_{\text{targ}, 1} \gets \phi_1$, $\phi_{\text{targ}, 2} \gets \phi_2$
\State \textbf{for} $t = 1$ to $T$ 
    \State \qquad $s_t \gets \texttt{env}.\text{get\_current\_state()}$
    \State \qquad $a_t \sim \pi_{\theta}(\cdot \mid s_t, t)$ 
    \State \qquad $(s_{t+1}, r_t) \gets \texttt{env}.\text{step}(a_t)$
    \State \qquad Store $(s_t, a_t, r_t, t)$ in $\mathcal{D}$
    \State \qquad \textbf{if} { $|\mathcal{D}| \geq B$}
        \State \qquad \qquad Randomly sample a batch $\mc{B} = \{ (s_t, a_t, r_t, t) \}$ from $\mc{D}$
        \State  \qquad \qquad  Compute targets for the critic networks:
        \[
            y(r_t, s_{t+1}) = r_t + \gamma \left( \min_{i = 1, 2} Q_{\phi_{\text{targ}, i}}(s_{t+1}, a', t+1) - \alpha \log \pi_{\theta} (a' \mid s_{t+1}, t+1) \right); \ a' \sim \pi_{\theta}(\cdot \mid s_{t+1}, t+1)
            \]
        \State  \qquad \qquad  Update the critic networks by one step of gradient descent using
            \[
            \nabla_{\phi_{\text{targ}, i}} \frac{1}{B} \sum_{(s_t, a_t, r_t, t) \in \mc{B}} \left( Q_{\phi_{\text{targ}, i}}(s_t, a_t, t) - y(r_t, s_{t+1}) \right)^2; \ \text{for} \ i=1,2
            \]
        \State  \qquad \qquad  Update policy network by one step of gradient ascent using
            \[
            \nabla_{\theta} \frac{1}{B} \sum_{s_t \in \mc{B}} \left( \min_{i = 1, 2} Q_{\phi_{\text{targ}, i}}(s_t, a_{\theta}(s_t), t) - \alpha \log \pi_{\theta} (a_{\theta}(s_t) \mid s_t, t) \right); \ a_{\theta}(s_t) \sim \pi_{\theta}(\cdot \mid s_t, t)
            \] 
        \State  \qquad \qquad  Update the target networks with
            \[
                \phi_{\text{targ}, i} \gets \rho \phi_{\text{targ}, i} + (1-\rho) \phi_{i}; \ \text{for} \ i=1,2
            \]
    \State  \qquad \textbf{end if}
\State \textbf{end for}
\end{algorithmic}
\end{algorithm}

\subsection{Proximal Policy Optimization}
\label{ssec:ppo}

Proximal Policy Optimization (PPO)~\citep{schulman2017ppo}  is an on-policy reinforcement learning algorithm that seeks to improve an agent's policy while ensuring that updates do not deviate too far from the current behavior.

While standard PPO is typically applied to episodic tasks with frequent environment resets, our implementation is adapted for a single-episode, reset-free regime. In this setting, the agent must continuously learn and adapt over a single, indefinite trajectory. We introduce several key modifications to the original framework:

\paragraph{Time-aware policy and value networks} To handle the non-stationary nature of the foraging task, we make both the policy ($\pi_\theta$) and critic ($V_\phi$) functions of time $t$. 

\paragraph{Infinite-Horizon Bootstrapping:} In the absence of terminal states, we modify the Generalized Advantage Estimation (GAE). The advantage $\hat{A}_t$ is calculated by bootstrapping from the critic’s value estimate of the next state for every transition, treating the simulation as a continuous stream of experience rather than a series of independent episodes.

\paragraph{Online transition management} The Rollout Buffer is utilized to store a fixed window of transitions. Updates are performed at regular intervals (every 512 steps), allowing for online policy refinement without the need for a global reset.

\noindent The online training loop of our PPO learner is outlined in~\cref{alg:online-ppo}.

\begin{algorithm}[H]
\caption{Online PPO for single-episode reset-free tasks}
\label{alg:online-ppo}
\begin{algorithmic}[1]
\State \textbf{Initialize:} Policy $\pi_\theta$, Value network $V_\phi$, Rollout Buffer $\mathcal{B}$
\State \textbf{Set parameters:} Clip $\epsilon$, GAE parameters $\gamma, \lambda$, update interval $N$, epochs $K$
\State \textbf{for} $t = 0, 1, 2, \dots, T$
    \State \qquad $s_t \gets \texttt{env}.\text{get\_current\_state()}$
    \State \qquad $a_t \sim \pi_{\theta}(\cdot \mid s_t, t)$ 
    \State \qquad $(s_{t+1}, r_t) \gets \texttt{env}.\text{step}(a_t)$
    \State \qquad Store $\left(s_t, a_t, t, r_t, \log\pi_\theta(a_t|s_t, t), V_\phi(s_t, t)\right)$  in $\mathcal{B}$

    \State \qquad \textbf{if} $(t + 1) \bmod N = 0$

         \State \qquad \qquad Bootstrap next value: $\hat{V}_{next} = V_\phi(s_{t+1}, t)$
         \State \qquad \qquad Estimate the advantages $\hat{A}_i$ using GAE
         \State \qquad \qquad Compute targets $R_i = \hat{A}_i + V_\phi(s_i, E_i)$ and normalize $\hat{A}$

         \State \qquad \qquad \textbf{for} $k = 1, \dots, K$
         
         \State \qquad \qquad \qquad Calculate ratio $r_i(\theta) = \exp(\log\pi_\theta(a_i|s_i, t) - \log\pi_{old}(a_i|s_i, t))$
         \State \qquad \qquad \qquad $L_{clip} = \mathbb{E}_i [\min(r_i(\theta)\hat{A}_i, \text{clip}(r_i(\theta), 1-\epsilon, 1+\epsilon)\hat{A}_i)]$
         \State \qquad \qquad \qquad $L_{v} = \mathbb{E}_i [(V_\phi(s_i, t) - R_i)^2]$
         \State \qquad \qquad \qquad $L_{ent} = \mathbb{E}_i [\text{Entropy}(\pi_\theta(\cdot|s_i, t))]$
         \State \qquad \qquad \qquad Update $\theta, \phi$ via Adam $\nabla_{\theta, \phi} (L^{clip} - c_1 L^{v} + c_2 L^{ent})$

         \State \qquad \qquad \textbf{end for}

        \State \qquad \qquad Clear buffer $\mathcal{B}$

    \State \qquad \textbf{end if}
\State \textbf{end for}
\end{algorithmic}
\end{algorithm}
\section{Stochastic Sampling}
\label{sec:sampling}

In~\Cref{s:algorithm_l4dc}, we show that during the online update stage, the next state is sampled using Boltzmann sampling~\citep{sutton1990integrated, cesa2017boltzmann}, where the selection probability of each candidate next state is proportional to its estimated prospective reward. Here, we compare two online PLuC variants: one using deterministic argmax style selection, and the other using Boltzmann sampling.

\begin{figure}[!t]
    \centering
    \includegraphics[width=0.8\linewidth]{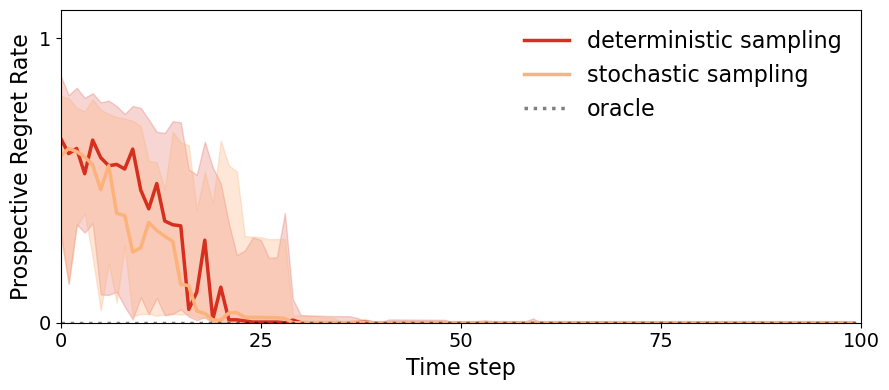}
    \caption{\textbf{Online PLuC with stochastic sampling and deterministic sampling both approach oracle level performance.} In this environment, both approaches reduce prospective regret over time and eventually approach oracle-level performance, while stochastic sampling (red) exhibits a slightly faster decrease at the beginning than the deterministic sampling (orange).}
    \label{f:sampling}
\end{figure}


\end{document}